\documentclass{article} % For LaTeX2e
\usepackage{iclr2015,times}
\usepackage{hyperref}
\usepackage{url}
\usepackage{algcompatible}
\usepackage{algorithm}
\usepackage{amsmath}
\usepackage{graphicx}
\usepackage{subfigure}
\usepackage{caption}

\title{FitNets: Hints for Thin Deep Nets}

\author{
Adriana Romero$^1$, Nicolas Ballas$^2$, Samira Ebrahimi Kahou$^3$, Antoine Chassang$^2$,\\
{\bf Carlo Gatta$^4$} \& {\bf Yoshua Bengio$^2\dagger$}\\
$^1$Universitat de Barcelona, Barcelona, Spain. \\
$^2$Universit\'e de Montr\'eal, Montr\'eal, Qu\'ebec, Canada. $^\dagger$CIFAR Senior Fellow.\\
$^3$\'Ecole Polytechnique de Montr\'eal, Montr\'eal, Qu\'ebec, Canada. \\
$^4$Centre de Visi\'o per Computador, Bellaterra, Spain.
}

\DeclareMathOperator*{\argmin}{argmin}
\newcommand*{\argminl}{\argmin\limits}

\iclrfinalcopy % Uncomment for camera-ready version

\iclrconference % Uncomment if submitted as conference paper instead of workshop

\begin{document}

\maketitle

%%% FIXME Put more emphasis on compression
\begin{abstract}
While depth tends to improve network performances, it also makes
gradient-based training more difficult since deeper networks tend to be more
non-linear. The recently proposed knowledge distillation approach is aimed at obtaining small
and fast-to-execute models, and it has shown that a student network could imitate
the soft output of a larger teacher network or ensemble of networks.  In this
paper, we extend this idea to allow the training of a student that is deeper and
thinner than the teacher, using not only the outputs but also the intermediate
representations learned by the teacher as hints to improve the training process
and final performance of
the student. Because the student intermediate hidden layer will generally be
smaller than the teacher's intermediate hidden layer, additional parameters
are introduced to map the student hidden layer to the prediction of the
teacher hidden layer. This allows one to train deeper students that can
generalize better or run faster, a trade-off that is controlled by the chosen
student capacity. For example, on CIFAR-10, a deep student network
with almost 10.4 times less parameters outperforms a larger,
state-of-the-art teacher network.
%Advantage of this formulation is two-folds, First,
%we are able to assess the difficulty of the training examples
%through the teacher and weight their importance in the model cost accordingly to ease the student optimization problem. In addition, teacher hidden-layers can also serve as intermediate hints which regularizes the training process to obtain deep students.

% FIXME put some number Number
%Our proposition is evaluated on MNIST, CIFAR-10, CIFAR-100 and SVHN datasets.
%Most importantly, we demonstrate using our training method our ability to learn deep and thin networks that achieve state-of-art results while keeping low the number of parameters and therefore the computational requirements.

\end{abstract}

\vspace*{-1mm}
\section{Introduction}
\vspace*{-1mm}

%%  Deep nets and soa & drawback (wide and deep)
Deep networks have recently exhibited state-of-the-art performance in computer vision tasks
such as image classification and object detection
\citep{simonyan.2014,Szegedy14}. However, top-performing systems usually involve very \emph{wide} and
\emph{deep} networks, with numerous parameters. Once learned, a major drawback
of such wide and deep models is that they result in very time consuming
systems at inference time, since they need to perform a huge number of
multiplications. Moreover, having large amounts of parameters makes the models
high memory demanding. For these reasons, wide and deep top-performing
networks are not well suited for applications with memory or time limitations.

%% Shallow vs deep & deep nets are hard to train
There have been several attempts in the literature to tackle the problem of model compression to reduce the computational burden at inference
time. In~\citet{Bucila06}, authors propose to train a neural network to mimic
the output of a complex and large ensemble. The method uses the ensemble to
label unlabeled data and trains the neural network with the data labeled by
the ensemble, thus mimicking the function learned by the ensemble and
achieving similar accuracy. The idea has been recently adopted in
\citet{ba.2013} to compress deep and wide networks into shallower but even
wider ones, where the compressed model mimics the function learned by the
complex model, in this case, by using data labeled by a deep (or an ensemble of
deep) networks. More recently, Knowledge Distillation (KD)~\citep{Hinton14} was introduced as a
model compression framework, which eases the training of deep networks by following a student-teacher
paradigm, in which the student is penalized according to a softened version of the teacher's
output. The framework compresses an ensemble of deep
networks (\emph{teacher}) into a \emph{student} network of \emph{similar depth}. To do so, the
student is trained to predict the output of the teacher, as
well as the true classification labels. All previous works related to Convolutional Neural Networks focus on compressing a teacher network or an ensemble of networks into either networks of similar width and depth or into shallower and wider ones; not taking advantage of depth.

%which does not trade depth for compression. The
%method eases the training of deep networks by following a student-teacher
%paradigm, in which the student is penalized according to the teacher's
%prediction. The framework compresses an ensemble (\emph{teacher}) of deep
%networks into a \emph{student} network of \emph{similar depth}. To do so, the
%student network is trained to predict the output of the teacher ensemble, as
%well as the true classification labels.

\emph{Depth} is a fundamental aspect of
representation learning, since it encourages the re-use of features, and leads
to more abstract and invariant representations at higher layers~\citep{bengio.2013}. The importance of depth has been verified (1)
theoretically: deep representations are exponentially more expressive than
shallow ones for some families of functions~\citep{Montufar14};
and (2) empirically: the two top-performers of ImageNet use deep convolutional
networks with 19 and 22 layers, respectively~\citep{simonyan.2014} and \citep{Szegedy14}.

%% Training deep nets
Nevertheless, training deep architectures has proven to be
challenging~\citep{Larochelle07, Erhan09}, since they are composed of
successive non-linearities and, thus result in highly non-convex and
non-linear functions. Significant effort has been devoted to alleviate
this optimization problem. On the one hand, pre-training strategies, whether
unsupervised~\citep{Hinton06_NC,Bengio07} or supervised~\citep{Bengio07} train the network parameters in a greedy
layerwise fashion in order to initialize the network parameters in a
potentially good basin of attraction. The layers are trained one after the
other according to an intermediate target. Similarly, semi-supervised embedding~\citep{WestonJ2008-small}
provides guidance to an intermediate layer to help learn very deep networks. Along this line of reasoning, \citep{Cho12} ease the optimization problem of DBM by borrowing the activations of another model every second layer in a purely unsupervised scenario.
More recently, \citep{Lee14,Szegedy14,Gulcehre13} showed that adding
supervision to intermediate layers of deep architectures assists the training
of deep networks. Supervision is introduced by stacking a supervised MLP with a softmax layer
on top of intermediate hidden layers to ensure their discriminability
w.r.t. labels. Alternatively, Curriculum Learning
strategies (CL)~\citep{bengio.2009} tackle the optimization problem
by modifying the training distribution, such that the learner
network gradually receives examples of increasing and appropriate difficulty w.r.t. the already learned concepts. As a result, curriculum learning acts like
a continuation method, speeds up the convergence of the training process and finds potentially better
local minima of highly non-convex cost functions.

%% What we do
In this paper, we aim to address the network compression problem by taking
advantage of depth. We propose a novel approach to train \emph{thin} and \emph{deep} networks, called \emph{FitNets}, to compress  \emph{wide} and shallower (but still \emph{deep}) networks. The method
is rooted in the recently proposed Knowledge Distillation (KD) \citep{Hinton14} and extends the
idea to allow for thinner and deeper student models. We introduce
\emph{intermediate-level hints} from the teacher hidden layers to guide the training process of the student,
\textit{i.e.}, we want the student network (FitNet) to learn an intermediate representation
that is predictive of the intermediate representations of the teacher network. Hints allow the training
of thinner and deeper networks. Results
confirm that having deeper models allow us to generalize better, whereas
making these models thin help us reduce the computational burden
significantly.  We validate the proposed method on MNIST, CIFAR-10, CIFAR-100, SVHN and AFLW benchmark datasets and provide evidence that our method matches or
outperforms the teacher's performance, while requiring notably fewer
parameters and multiplications.

\vspace*{-1mm}
\section{Method}
\label{sec:method}
\vspace*{-1mm}

In this section, we detail the proposed student-teacher framework to train FitNets from shallower and wider nets. First, we review the recently
proposed KD. Second, we highlight the proposed hints algorithm to guide the
FitNet throughout the training process. Finally, we describe how the
FitNet is trained in a stage-wise fashion.

\vspace*{-1mm}
\subsection{Review of Knowledge Distillation}
\vspace*{-1mm}

%% Top layer training
In order to obtain a faster inference, we explore the recently proposed compression framework \citep{Hinton14}, which trains a \textit{student
  network}, from the softened output of an ensemble of wider networks,
\textit{teacher network}. The idea is to allow the student network to capture
not only the information provided by the true labels, but also the finer structure
learned by the teacher network. The framework can be
summarized as follows.

Let $\mathrm{T}$ be a teacher network with an output softmax $\mathrm{P_T}={\rm softmax}(\mathbf{a}_T)$
where $\mathbf{a}_T$ is the vector of teacher pre-softmax activations, for some example. In the case where the teacher model
is a single network, $\mathbf{a}_T$ represents the weighted sums of the output layer, whereas if
the teacher model is the result of an ensemble either $\mathrm{P_T}$ or $\mathbf{a}_T$ are obtained
by averaging outputs from different networks (respectively for arithmetic or geometric averaging).
Let $\mathrm{S}$ be a student network with parameters $\mathbf{W_S}$ and output probability $\mathrm{P_S} = \mathrm{softmax}(\mathbf{a}_S)$, where $\mathbf{a}_S$ is the
%YB% and note that P(Y|X) notation was not correctly used so I removed it, it cannot be a VECTOR of probabilities for ALL Y's, but the probability for a single Y.
student's pre-softmax output. The student network will be trained
such that its output $\mathrm{P_S}$ is similar to the teacher's output
$\mathrm{P_T}$, as well as to the true labels $\mathbf{y_{true}}$. Since
$\mathrm{P_T}$ might be very close to the one hot code representation of the
sample's true label, a relaxation $\tau > 1$ is introduced to soften the signal
arising from the output of the teacher network, and thus, provide more information during
training\footnote{For example, as argued by~\citet{Hinton14}, with softened outputs, more information is provided about the
relative similarity of the input to classes other than the one with the highest probability.}. 
The same relaxation is applied to the output of the student network ($\mathrm{P_S^\tau}$), 
when it is compared to the teacher's softened output ($\mathrm{P_T^\tau}$):
%=======
%sample's true label, a relaxation $\tau > 1$ is introduced to soften the
%output of the teacher network:
%>>>>>>> .r24636
\begin{equation}
\mathrm{P_T^\tau} = \mathrm{softmax}\left(\frac{\mathbf{a}_T}{\tau}\right),\quad \mathrm{P_S^\tau} = \mathrm{softmax}\left(\frac{\mathbf{a}_S}{\tau}\right).
\end{equation}

%\begin{equation}
%\mathrm{P_S^\tau} = \mathrm{softmax}(\frac{\mathbf{a}_S}{\tau}).
%\end{equation}

The student network is then trained to optimize the following loss
function:

\begin{equation}
\mathcal{L}_{KD}(\mathbf{W_S}) = \mathcal{H}(\mathbf{y_{true}},\mathrm{P_S}) +
\lambda \mathcal{H}(\mathrm{P_T^\tau}, \mathrm{P_S^\tau}),
\label{eq:toplayer}
\end{equation}
where $\mathcal{H}$ refers to the cross-entropy and $\lambda$ is a
tunable parameter to balance both cross-entropies. Note that the first term in
Eq. (\ref{eq:toplayer}) corresponds to the traditional cross-entropy between the
output of a (student) network and labels, whereas the second term enforces the
student network to learn from the softened output of the teacher network.

%% light transition
To the best of our knowledge, KD is designed such that student networks mimic
teacher architectures of similar depth. Although we found the KD framework to achieve
encouraging results even when student networks have slightly deeper
architectures, as we increase the depth of the student network, KD training
still suffers from the difficulty of optimizing deep nets
(see Section \ref{ssec:benefits}).

\vspace*{-1mm}
\subsection{Hint-based Training}
\vspace*{-1mm}

%% hints training
In order to help the training of deep FitNets (deeper than their teacher), we
%=======
%In order to train FitNets, we
%>>>>>>> .r24636
introduce \textit{hints} from the teacher network. A \textit{hint} is defined
as the output of a teacher's hidden layer responsible for guiding the
student's learning process.  Analogously, we choose a hidden
layer of the FitNet, the \textit{guided} layer, to learn from the teacher's hint layer. We want
%YB% it does not make sense to call the guided layer that receive a hint the ``target'' layer, so I renamed
%YB% everywhere to ``guided'' or ``guided layer''. Please change the figures accordingly.
the guided layer to be able to predict the output of the hint layer. Note that having
hints is a form of regularization and thus, the pair hint/guided layer has to be
chosen such that the student network is not over-regularized. The deeper we set the guided layer, the less flexibility we give to the network and, therefore, FitNets are more likely to suffer from over-regularization. In our case, we
%YB% I removed reference to convolutions where it was not necessary, because the
%YB% proposed approach is not limited to convnets
choose the hint to be the middle layer of the teacher
network. Similarly, we choose the guided layer to be the middle layer
%=======
%chosen, such that the student network is not over-regularized. The deeper we set the target layer, the less flexibility we give to the network and, therefore the FitNet is more likely to suffer from over-regularization. In our case, we choose the hint to be the middle convolutional layer of the teacher
%network. Similarly, we choose the target to be the middle convolutional layer
%>>>>>>> .r24636
of the student network. 

Given that the teacher network will usually be wider than the FitNet,
the selected hint layer may have more outputs than the guided layer. For that
reason, we add a regressor to the guided layer, whose output matches the size of the
hint layer. Then, we train the FitNet parameters from the first layer up to
the guided layer as well as the regressor parameters by minimizing the
following loss function:
%Maybe it would be good to have a figure to better explain it.
\begin{equation}
\mathcal{L}_{HT}(\mathbf{W_{Guided}},\mathbf{W_r}) = \frac{1}{2} || u_{h}(\mathbf{x};\mathbf{W_{Hint}}) - r(v_{g}(\mathbf{x};\mathbf{W_{Guided}});\mathbf{W_r}) ||^2,
\label{eq:hints}
\end{equation}

where $u_{h}$ and $v_{g}$ are the teacher/student deep nested functions up to
their respective hint/guided layers with parameters $\mathbf{W_{Hint}}$ and
$\mathbf{W_{Guided}}$, $r$ is the regressor function on top of the guided
layer with parameters $\mathbf{W_r}$. Note that the outputs of $u_{h}$ and $r$
have to be comparable, \textit{i.e.}, $u_{h}$ and $r$ must be the same non-linearity.

Nevertheless, using a fully-connected regressor increases the number of
parameters and the memory consumption dramatically in the case
where the guided and hint layers are convolutional. Let $\mathrm{N}_{h,1}
\times \mathrm{N}_{h,2}$ and $\mathrm{O}_{h}$ be the teacher hint's spatial
size and number of channels, respectively. Similarity, let $\mathrm{N}_{g,1}
\times \mathrm{N}_{g,2}$ and $\mathrm{O}_{g}$ be the FitNet guided layer's spatial
size and number of channels. The number of parameters in the weight matrix of
a fully connected regressor is $\mathrm{N}_{h,1} \times \mathrm{N}_{h,2}
\times \mathrm{O}_{h} \times \mathrm{N}_{g,1} \times \mathrm{N}_{g,2} \times
\mathrm{O}_{g}$. To mitigate this limitation, we use a convolutional regressor
instead. The convolutional regressor is designed such that it considers
approximately the same spatial region of the input image as the teacher
hint. Therefore, the output of the regressor has the same spatial size as the
teacher hint. Given a teacher hint of spatial size $\mathrm{N}_{h,1} \times
\mathrm{N}_{h,2}$, the regressor takes the output of the Fitnet's guided
layer of size $\mathrm{N}_{g,1} \times \mathrm{N}_{g,2}$ and adapts its kernel
shape $\mathrm{k_1} \times \mathrm{k_2}$ such that $\mathrm{N}_{g,i} -
\mathrm{k}_i + 1 = \mathrm{N}_{h,i}$, where $i \in \{1,2\}$. The number of
parameters in the weight matrix of a the convolutional regressor is
$\mathrm{k_1} \times \mathrm{k_2} \times \mathrm{O}_{h} \times
\mathrm{O}_{g}$, where $\mathrm{k_1} \times \mathrm{k_2}$ is significantly
lower than $\mathrm{N}_{h,1} \times \mathrm{N}_{h,2} \times \mathrm{N}_{g,1}
\times \mathrm{N}_{g,2}$.

\vspace*{-1mm}
\subsection{FitNet Stage-wise Training}
\vspace*{-1mm}

\begin{figure}[t]
\centering
\subfigure[Teacher and Student Networks]{\includegraphics[width=0.32\textwidth]{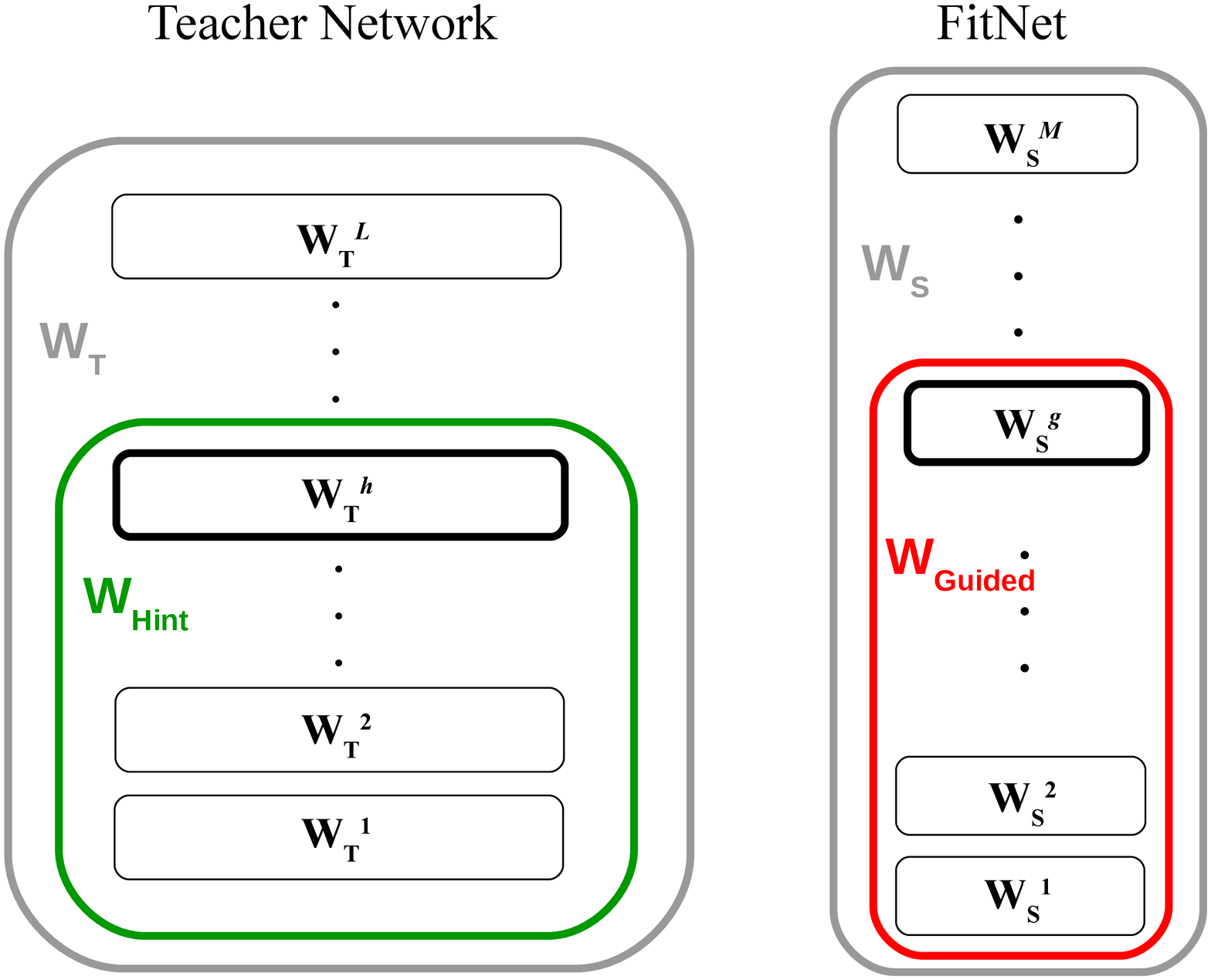}}
\subfigure[Hints Training]{\includegraphics[width=0.32\textwidth]{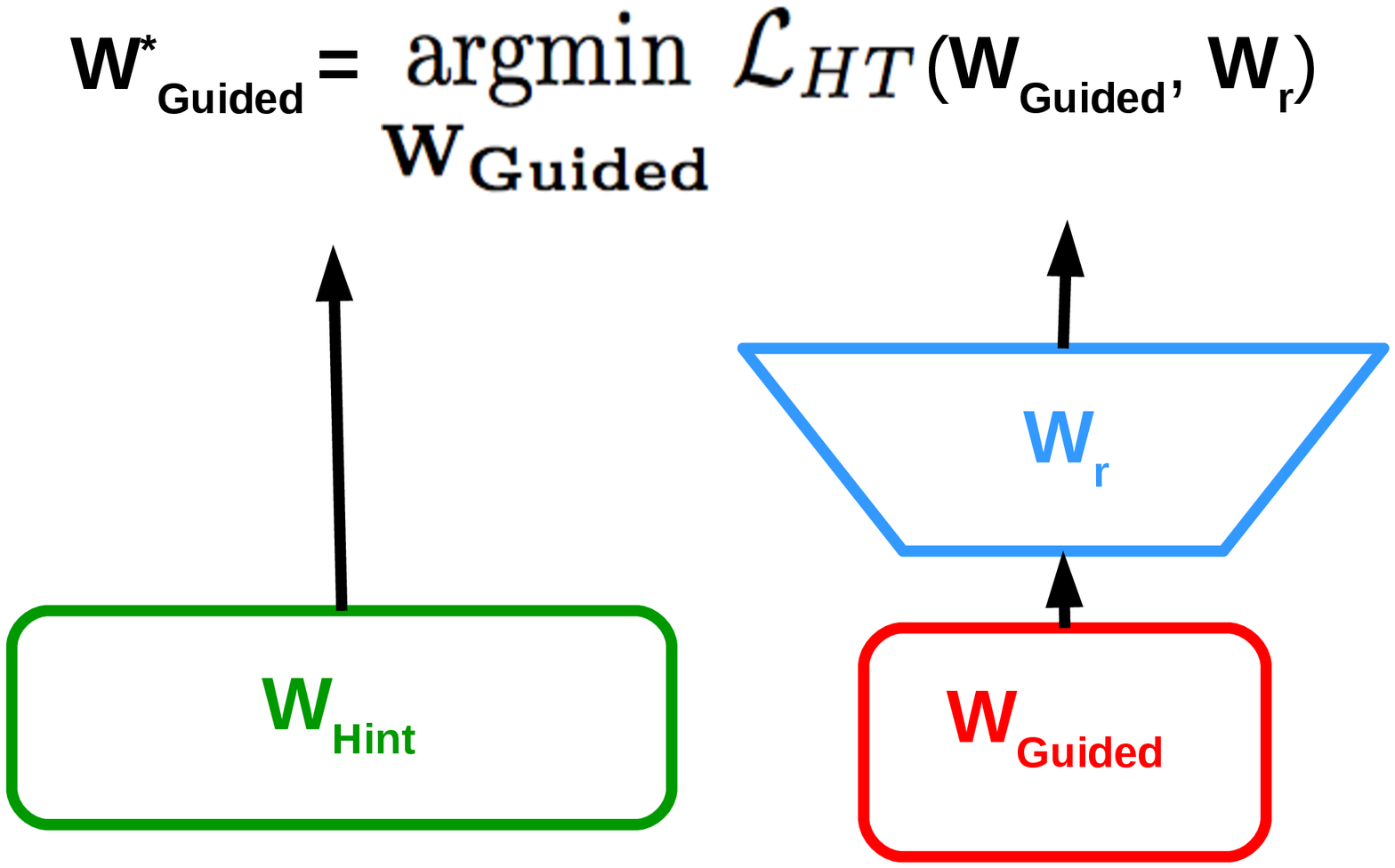} \label{fig:hints}}
\subfigure[Knowledge Distillation]{\includegraphics[width=0.32\textwidth]{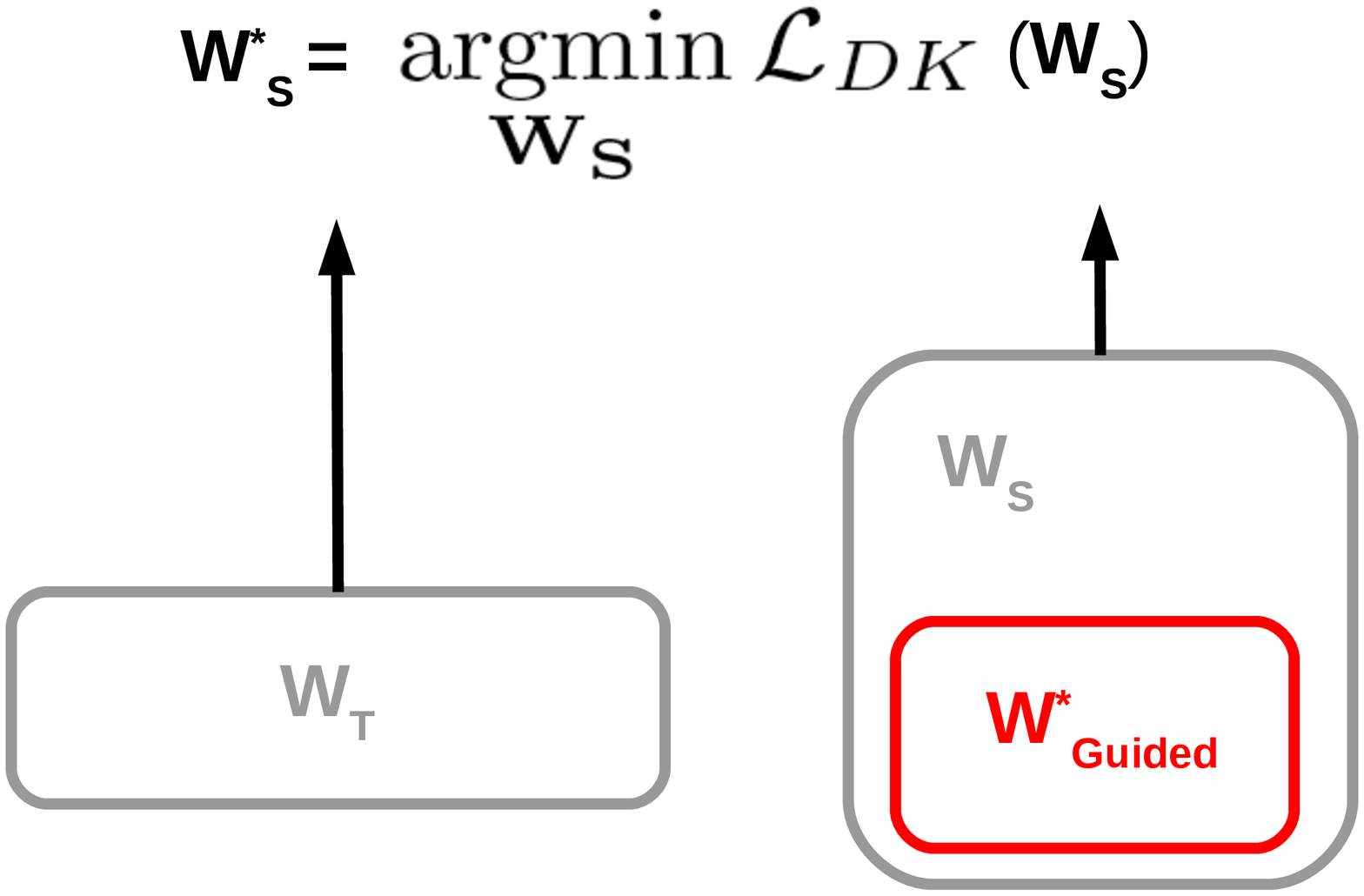} \label{fig:top_layer}}
\caption{Training a student network using hints.}\label{fig:method}
\end{figure}

%YB% It is not layerwise because we do not do this for every layer
%YB% but it is stage-wise because we have 2 stages, pre-training up to guided layer, then fine-tuning by KD with annealing
We train the FitNet in a stage-wise fashion following the student/teacher
paradigm. Figure \ref{fig:method} summarizes the training pipeline.
Starting from a trained teacher network and a randomly initialized FitNet
(Fig. \ref{fig:method} (a)), we add a regressor parameterized by
$\mathbf{W_{r}}$ on top of the FitNet guided layer and train the FitNet
parameters $\mathbf{W_{Guided}}$ up to the guided layer to minimize
Eq. (\ref{eq:hints}) (see Fig. \ref{fig:method} (b)).  Finally, from the
pre-trained parameters, we train the parameters of whole FitNet
$\mathbf{W_{S}}$ to minimize Eq. (\ref{eq:toplayer}) (see
Fig. \ref{fig:method} (c)). Algorithm \ref{alg:StudentTraining} details the
FitNet training process.

% Starting from a trained teacher network with parameters
%$\mathbf{W_T}$, a FitNet with randomly initialized parameters
%$\mathbf{W_S}$, and two indices $h$ and $t$ corresponding to the teacher hint
%and the FitNet guided layers respectively, the algorithm proceeds as
%follows. The set of parameters $\mathbf{W_{Hint}}$ is defined as the teacher's
%parameters up to the hint layer $h$ (line 1). The set of parameters
%$\mathbf{W_{Guided}}$ is defined as the FitNet's parameters up to the guided
%layer $t$ (line 2). The regressor's parameters
%$\mathbf{W_r}$ are initialized to small random values (line 3). After that,
%the parameters $\mathbf{W_{Guided}}$ are trained by means of backprop to
%minimize Eq. (\ref{eq:hints} )(line 4) and, subsequently used to initialize the
%FitNet's parameters up to the guided layer, throwing away the regressor (line
%5). Finally, the parameters $\mathbf{W_{S}}$ of the whole student network are
%trained by means of backprop to minimize Eq. (\ref{eq:toplayer}) (line 6).

\begin{algorithm}
\caption{FitNet Stage-Wise Training.\newline 
  The algorithm receives as
  input the trained parameters $\mathbf{W_T}$ of a teacher, the randomly
  initialized parameters $\mathbf{W_S}$ of a FitNet, and two indices $h$
  and $g$ corresponding to hint/guided layers, respectively. Let
  $\mathbf{W_{Hint}}$ be the teacher's parameters up to the hint layer
  $h$. Let $\mathbf{W_{Guided}}$ be the FitNet's parameters up to the
  guided layer $g$. Let $\mathbf{W_r}$ be the regressor's parameters. The first
  stage consists in pre-training the student network up to the guided
  layer, based on the prediction error of the teacher's hint layer (line
  4). The second stage is a KD training of the whole network (line 6).
}
\label{alg:StudentTraining}
\begin{algorithmic}[1]
% \scriptsize
\small
\STATEx \textbf{Input:} $\mathbf{W_S}, \mathbf{W_T}, g, h$
\STATEx \textbf{Output:} $\mathbf{W_S^*}$

\STATE $\mathbf{W_{Hint}} \leftarrow  \{\mathbf{W_T}^1, \ldots, \mathbf{W_T}^h\}$
\STATE $\mathbf{W_{Guided}} \leftarrow  \{\mathbf{W_S}^1, \ldots, \mathbf{W_S}^g\}$

%\IF{$\mathrm{O_h} > \mathrm{O_t}$}
\STATE Intialize $\mathbf{W_r} $ to small random values
%\STATE $\mathbf{W_{Guided}} \leftarrow \mathbf{W_{Guided}} \cup \mathbf{W_r}$
%\ENDIF

\STATE $\mathbf{W_{Guided}^*} \leftarrow \argminl_\mathbf{W_{Guided}} \mathcal{L}_{HT}(\mathbf{W_{Guided}},\mathbf{W_r})$

%\STATE Train $\mathbf{W_{Guided}}$  using backprop and $\mathbf{W_{Hints}}$ to minimize Eq. \ref{eq:hints}

\STATE $\{\mathbf{W_S}^1, \ldots, \mathbf{W_S}^g\} \leftarrow \{\mathbf{W_{Guided}}^{*1}, \ldots, \mathbf{W_{Guided}}^{*g}\}$

\STATE $\mathbf{W_S^*} \leftarrow \argminl_\mathbf{W_S} \mathcal{L}_{KD}(\mathbf{W_S})$

%\STATE Train $\mathbf{W_{S}}$ using backprop and $\mathbf{W_T}$ to minimize Eq. \ref{eq:toplayer}

\end{algorithmic}
\vspace*{-1mm}
\end{algorithm}

\vspace*{-2mm}
\subsection{Relation to Curriculum Learning}
\vspace*{-1mm}

%YB% I prefer to spell out curriculum learning and not abbreviate it
In this section, we argue that our hint-based training with KD can be seen as a particular form of Curriculum Learning~\citep{bengio.2009}.  Curriculum learning has proven to accelerate
the training convergence as well as potentially improve the model
generalization by properly choosing a sequence of training distributions seen by the
learner: from simple examples to more complex ones. A
curriculum learning extension~\citep{Gulcehre13} has also shown that by using
guidance hints on an intermediate layer during the training, one could considerably ease training.
  However, ~\cite{bengio.2009} uses hand-defined
heuristics to measure the ``simplicity'' of an example in a sequence and~\cite{Gulcehre13}'s guidance hints require some prior knowledge of the
end-task. Both of these curriculum learning strategies tend to be problem-specific.

Our approach alleviates this issue by using a teacher model.  Indeed,
intermediate representations learned by the teacher are used as hints to guide
the FitNet optimization procedure. In addition, the teacher confidence
provides a measure of example ``simplicity'' by means of teacher cross-entropy
term in Eq. (\ref{eq:toplayer}). This term ensures that examples with a high
teacher confidence have a stronger impact than examples with low teacher
confidence: the latter correspond to probabilities closer to the uniform distribution,
which exert less of a push on the student parameters. In other words,
the teacher penalizes the training examples
according to its confidence. Note that parameter $\lambda$ in
Eq. (\ref{eq:toplayer}) controls the weight given to the teacher
cross-entropy, and thus, the importance given to each example. In order to
promote the learning of more complex examples (examples with lower teacher
confidence), we gradually anneal $\lambda$ during the training with a linear
decay. The curriculum can be seen as composed of two stages: first
learn intermediate concepts via the hint/guided layer transfer, then train the whole
student network jointly, annealing 
$\lambda$, which allows easier examples (on which the teacher is very confident)
to initially have a stronger effect, but
progressively decreasing their importance as $\lambda$ decays.
Therefore, the hint-based training introduced in the paper is a
generic curriculum learning approach, where prior information about the task-at-hand is deduced
purely from the teacher model.

\begin{minipage}[b]{.45\textwidth}
\centering
\small
\begin{tabular}{|c|c|c|}
 \hline
 \textbf{Algorithm} & \textbf{$\#$ params}  & \textbf{Accuracy}\\ \hline
\hline
\multicolumn{3}{|l|}{\emph{Compression}} \\ \hline
FitNet  & $\sim$2.5M& $\mathbf{91.61\%}$  \\ \hline
Teacher & $\sim$9M& $90.18\%$ \\ \hline
Mimic single & $\sim$54M & $84.6\%$  \\ \hline 
Mimic single & $\sim$70M & $84.9\%$  \\ \hline 
Mimic ensemble & $\sim$70M & $85.8\%$  \\ \hline \hline 
\multicolumn{3}{|l|}{\emph{State-of-the-art methods}} \\ \hline
\multicolumn{2}{|l|}{Maxout} & $90.65\%$   \\ \hline
\multicolumn{2}{|l|}{Network in Network} & $91.2\%$  \\ \hline
\multicolumn{2}{|l|}{Deeply-Supervised Networks} & $\mathbf{91.78\%}$ \\ \hline
\multicolumn{2}{|l|}{Deeply-Supervised Networks (19)} & $88.2\%$  \\ \hline 
\end{tabular}
\captionof{table}{Accuracy on CIFAR-10}
\label{tab:cifar10}
\end{minipage}\qquad
\begin{minipage}[b]{.45\textwidth}
\small
\centering
\begin{tabular}{|c|c|c|}
\hline
\textbf{Algorithm} & \textbf{$\#$ params}  & \textbf{Accuracy}\\ \hline
\hline
\multicolumn{3}{|l|}{\emph{Compression}} \\ \hline
FitNet & $\sim$2.5M& $\mathbf{64.96\%}$  \\ \hline
Teacher & $\sim$9M& $63.54\%$ \\ \hline \hline
\multicolumn{3}{|l|}{\emph{State-of-the-art methods}} \\ \hline
\multicolumn{2}{|l|}{Maxout} & $61.43\%$   \\ \hline
\multicolumn{2}{|l|}{Network in Network} & $64.32\%$  \\ \hline
\multicolumn{2}{|l|}{Deeply-Supervised Networks} & $\mathbf{65.43\%}$ \\ \hline
\end{tabular}
\captionof{table}{Accuracy on CIFAR-100}
\label{tab:cifar100}
\end{minipage}

\vspace*{-1mm}
\section{Results on Benchmark Datasets}
\label{sec:exp}
\vspace*{-1mm}

In this section, we show the results on several benchmark datasets\footnote{Code to reproduce the experiments publicly available: https://github.com/adri-romsor/FitNets}. The architectures of all networks as well as the training details are reported in the supplementary material. 

\vspace*{-1mm}
\subsection{CIFAR-10 and CIFAR-100}
\vspace*{-1mm}

%% About dataset
The CIFAR-10 and CIFAR-100 datasets \citep{Krizhevsky09} are composed of 32x32 pixel RGB images belonging to 10 and 100 different classes, respectively. They both contain 50K training images and 10K test images. CIFAR-10 has 1000 samples per class, whereas CIFAR-100 has 100 samples per class. 
Like \citet{Goodfellow13}, we normalized the datasets for contrast normalization and applied ZCA whitening.

%% About exp CIFAR-10
\textbf{CIFAR-10}: To validate our approach, we trained a teacher network of maxout convolutional layers as reported in \citet{Goodfellow13} and designed a FitNet with
17 maxout convolutional layers, followed by a maxout fully-connected layer and a top softmax layer, with roughly $1/3$ of the
parameters. The 11th layer of the student network was trained to mimic the 2nd layer of the teacher network. 
Like in \citet{Goodfellow13,Lee14},  we augmented the data with random flipping during training. Table~\ref{tab:cifar10} summarizes the obtained results. Our student model
outperforms the teacher model, while requiring notably fewer parameters,
suggesting that depth is crucial to achieve better
representations. When compared to network compression methods, our algorithm
achieves outstanding results; \textit{i.e.}, the student network achieves an accuracy
of $91.61\%$, which is significantly higher than the top-performer $85.8\%$
of \cite{ba.2013}, while requiring roughly 28 times fewer
parameters. When compared to state-of-the-art methods, our
algorithm matches the best performers. 

One could argue the choice of hinting the inner layers with the hidden state of a wide teacher network. A straightforward alternative would be to hint them with the desired output. This could be addressed in a few different ways: (1) Stage-wise training, where stage 1 optimizes the 1st half of the network w.r.t. classification targets and stage 2 optimizes the whole network w.r.t. classification targets. In this case, stage 1 set the network parameters in a good local minima but such initialization did not seem to help stage 2 sufficiently, which failed to learn. To further assist the training of the thin and deep student network, we could add extra hints with the desired output at different hidden layers. Nevertheless, as observed in \citep{Bengio07}, with supervised pre-training the guided layer may discard some factors from the input, which require
more layers and non-linearity before they can be exploited to predict the classes. (2) Stage-wise training with KD, where stage 1 optimizes the 1st half of the net w.r.t. classification targets and stage 2 optimizes the whole network w.r.t. Eq. (\ref{eq:toplayer}). As in the previous case, stage 1 set the network parameters in a good local minima but such initialization did not seem to help stage 2 sufficiently, which failed to learn. (3) Jointly optimizing both stages w.r.t. the sum of the supervised hint for the guided layer and classification target for the output layer. We performed this experiment, tried different initializations and learning rates with RMSprop \citep{Tieleman12} but we could not find any combination to make the network learn. Note that we could ease the training by adding hints to each layer and optimizing jointly as in Deeply Supervised Networks (DSN). Therefore, we built the above-mentioned 19-layer architecture and trained it by means of DSN, achieving a test performance of $88.2\%$, which is significantly lower than the performance obtained by the FitNets hint-based training ($91.61\%$). Such result suggests that using a very discriminative hint w.r.t. classification at intermediate layers might be too aggressive; using a smoother hint (such as the guidance from a teacher network) offers better generalization. (4) Jointly optimizing both stages w.r.t. the sum of supervised hint for the guided layer and Eq. (\ref{eq:toplayer}) for the output layer. Adding supervised hints to the middle layer of the network did not ease the training of such a thin and deep network, which failed to learn.

%% About exp CIFAR-100
\textbf{CIFAR-100}: To validate our approach, we trained a teacher network of maxout convolutional layers as reported in \citet{Goodfellow13} and used the same FitNet architecture as in CIFAR-10. As in \citet{Lee14},  we augmented the data with random flipping during training. Table~\ref{tab:cifar100} summarizes the obtained results. As in the previous case, our FitNet outperforms the teacher model, 
reducing the number of parameters by a factor of 3 and, when compared to state-of-the-art methods, the FitNet provides near state-of-the-art performance.

\vspace*{-1mm}
\subsection{SVHN}
\vspace*{-1mm}

%% About dataset
The SVHN dataset~\citep{netzer.2011} is composed by $32 \times 32$ color images of house numbers collected by GoogleStreet View. There are 73,257 images in the training set, 26,032 images in the test set and 531,131 less difficult examples. We follow the evaluation procedure of \cite{Goodfellow13} and use their maxout network as teacher. We trained a 13-layer FitNet composed of 11 maxout convolutional layers, a fully-connected layer and a softmax layer.

%% About exp

\begin{minipage}[b]{.45\textwidth}
\small
\centering
\begin{tabular}{|c|c|c|}
\hline
\textbf{Algorithm} & \textbf{$\#$ params}  & \textbf{Misclass}\\ \hline
\hline
\multicolumn{3}{|l|}{\emph{Compression}} \\ \hline
FitNet & $\sim$1.5M &  $2.42\%$\\ \hline
Teacher & $\sim$4.9M & $\mathbf{2.38\%}$\\ \hline  \hline
\multicolumn{3}{|l|}{\emph{State-of-the-art methods}} \\ \hline
\multicolumn{2}{|l|}{Maxout}  & $2.47\%$\\ \hline
\multicolumn{2}{|l|}{Network in Network} &  $2.35\%$\\ \hline
\multicolumn{2}{|l|}{Deeply-Supervised Networks} &  $\mathbf{1.92\%}$\\ \hline
\end{tabular}
\captionof{table}{SVHN error}
\label{tab:svhn}
\end{minipage}\qquad
\begin{minipage}[b]{.45\textwidth}
\small
\centering
\begin{tabular}{|c|c|c|}
\hline
\textbf{Algorithm} & \textbf{$\#$ params}  & \textbf{Misclass}\\ \hline
\hline
\multicolumn{3}{|l|}{\emph{Compression}} \\ \hline
Teacher & $\sim$361K & $0.55\%$ \\ \hline
Standard backprop & $\sim$30K & $1.9\%$ \\ \hline
KD & $\sim$30K & $0.65\%$ \\ \hline
FitNet & $\sim$30K & $\mathbf{0.51\%}$ \\ \hline
\multicolumn{3}{|l|}{\emph{State-of-the-art methods}} \\ \hline \hline
\multicolumn{2}{|l|}{Maxout} & $0.45\%$ \\ \hline
\multicolumn{2}{|l|}{Network in Network} & $0.47\%$ \\ \hline
\multicolumn{2}{|l|}{Deeply-Supervised Networks} & $\mathbf{0.39\%}$ \\ \hline
\end{tabular}
\captionof{table}{MNIST error}
%\caption{Accuracy on CIFAR-100}
\label{tab:mnist}
\end{minipage}

Table~\ref{tab:svhn} shows that our FitNet achieves comparable accuracy than the teacher
despite using only $32\%$ of teacher capacity. Our FitNet is comparable in terms of performance
to other state-of-art methods, such as Maxout and Network in Network.

\vspace*{-1mm}
\subsection{MNIST}
\vspace*{-1mm}

%% About dataset
As a sanity check for the training procedure, we evaluated the proposed method
on the MNIST dataset \citep{Lecun98}. MNIST is a dataset of handwritten
digits (from 0 to 9) composed of 28x28 pixel greyscale images, with 60K
training images and 10K test images. We trained a teacher network of maxout convolutional layers as reported in
\citet{Goodfellow13} and designed a FitNet twice as deep as the teacher network and with roughly $8\%$ of the
parameters. The 4th layer of the student network was trained to mimic the 2nd layer of the teacher network.

Table \ref{tab:mnist} reports the obtained results. To verify the influence of
using hints, we trained the FitNet architecture using either (1) standard backprop (w.r.t. classification labels), (2) KD
or (3) Hint-based Training (HT). When training the FitNet with standard backprop from the softmax layer, the deep and thin architecture achieves 
$1.9\%$ misclassification error. Using KD, the very same network achieves $0.65\%$, which confirms the potential of the teacher network; 
and when adding hints, the error still decreases to $0.51\%$. Furthermore, the student network achieves slightly better
results than the teacher network, while requiring 12 times less parameters.

\vspace*{-1mm}
\subsection{AFLW}
\vspace*{-1mm}

%% About dataset
AFLW \citep{AFLW} is a real-world face database, containing 25K annotated images. In order to evaluate the proposed framework in a face recognition setting, we extracted positive samples by re-sizing the annotated regions of the images to fit 16x16 pixels patches. Similarly, we extracted 25K 16x16 pixels patches not containing faces from ImageNet \citep{ImageNet} dataset, as negative samples. We used  $90\%$ of the extracted patches to train the network.

%% About exp
In this experiment, we aimed to evaluate the method on a different kind of architecture. Therefore, we trained a teacher network of 3 ReLU convolutional layers and a sigmoid output layer. We designed a first FitNet (FitNet 1) with 15 times fewer multiplications than the teacher network, and a second FitNet (FitNet 2) with 2.5 times fewer multiplications than the teacher network. Both FitNets have 7 ReLU convolutional layers and a sigmoid output layer.

The teacher network achieved $4.21\%$ misclassification error on the validation set. We trained both FitNets by means of KD and HT. On the one hand, we report a misclassification error of $4.58\%$ when training FitNet 1 with KD and a misclassification error of when $2.55\%$ when training it with HT. On the other hand, we report a missclassifation error of $1.95\%$ when training FitNet 2 with KD and a misclassification error of $1.85\%$ when training it with HT. These results show how the method is extensible to different kind of architectures and highlight the benefits of using hints, especially when dealing with thinner architectures.

\vspace*{-1mm}
\section{Analysis of empirical results}
\label{sec:empres}
\vspace*{-1mm}

We empirically investigate the benefits of our
approach by comparing various networks trained using
standard backpropagation (cross-entropy w.r.t. labels), KD
or Hint-based Training (HT). Experiments are performed on CIFAR-10
dataset~\citep{Krizhevsky09}.

%% More details on the capacity of parameters
We compare networks of increasing depth given a fixed computational
budget. Each network is composed of successive convolutional layers of kernel
size $3 \times 3$, followed by a maxout non-linearity and a non-overlapping $2
\times 2$ max-pooling. The last max-pooling takes the maximum over all remaining
spatial dimensions leading to a $1 \times 1$ vector representation.  We only
change the depth and the number of channels per convolution between different
networks, \textit{i.e.} the number of channels per convolutional layer
decreases as a network depth increases to respect a given computational budget. 
% The detail of all the network architectures are provided in the supplementary material.

\vspace*{-1mm}
\subsection{Assisting the training of deep networks}
\label{ssec:benefits}
\vspace*{-1mm}

\begin{figure}
\centering
\subfigure[30M Multiplications]{
\label{fig:depthexpa}
\includegraphics[width=5.5cm]{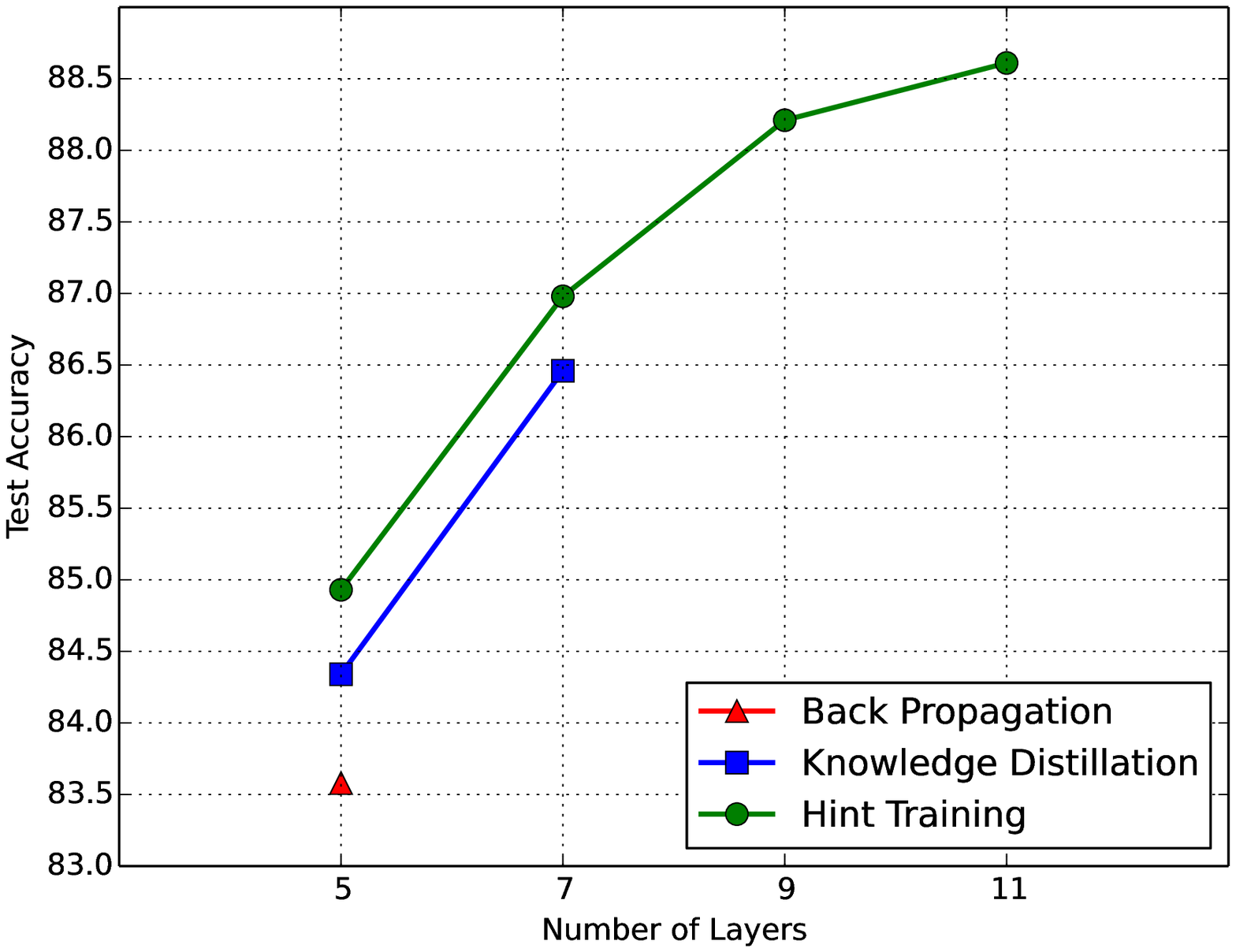}
}
\subfigure[107M Multiplications]{
\label{fig:depthexpb}
\includegraphics[width=5.5cm]{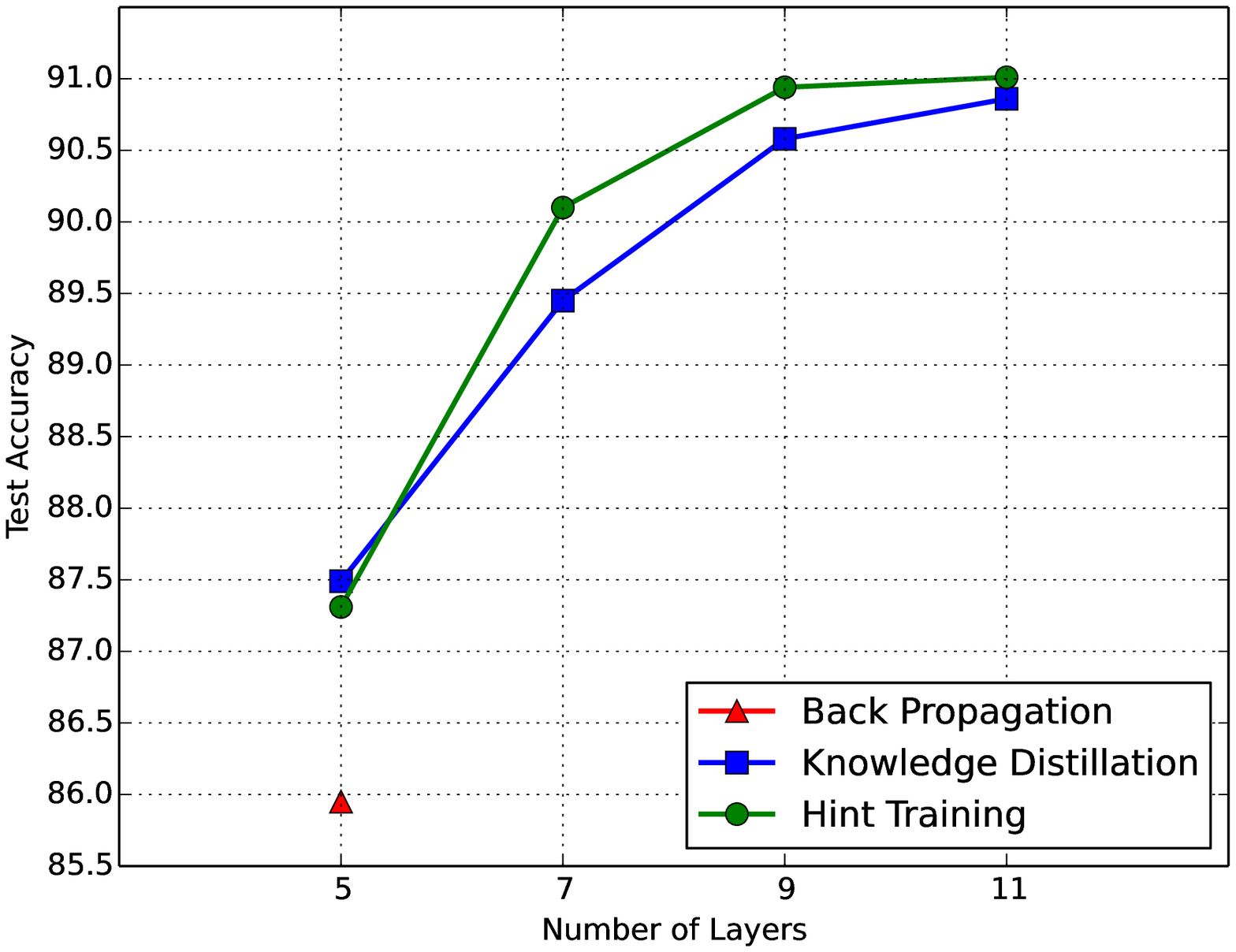}
}
\caption{Comparison of Standard Back-Propagation, Knowledge Distillation and Hint-based Training on CIFAR-10.}
\label{fig:depthexp}
\end{figure}
In this section, we investigate the impact of HT.  We
consider two computational budgets of approximately 30M and 107M
operations, corresponding to the multiplications needed in an image forward
propagation.
For each computational budget, we train networks composed of 3, 5, 7 and 9
convolutional layers, followed by a fully-connected layer and a softmax layer.
We compare their performances when they are trained with standard
backpropagation, KD and HT.  Figure~\ref{fig:depthexp} reports test on CIFAR-10 using early stopping on 
the validation set,
\textit{i.e.} we do not retrain our models on the training plus validation sets.

%% Optimization
Due to their depth and small capacity, FitNets are hard to train. As shown in Figure~\ref{fig:depthexpa}, we could
not train 30M multiplications networks with more than 5 layers with standard
backprop. When using KD, we succesfully trained networks up to 7
layers.
% It sucessfully trained train networks up to 7 layers.
Adding KD's teacher cross-entropy to the training objective (Eq. (\ref{eq:toplayer}))
gives more importance to easier examples, \textit{i.e.} samples for which the teacher
network is confident and, can lead to a smoother version of the training
cost~\citep{bengio.2009}. Despite some optimization benefits, it is worth
noticing that KD training still suffers from the increasing depth and reaches
its limits for 7-layer networks. HT tends to ease these optimization issues and is able to train 13-layer networks of 30M multiplications.
 The only difference between HT and KD is the starting point
in the parameter space: either random or obtained by means of the teacher's
hint. On the one hand, the proliferation of local minima and especially saddle points in
highly non-linear functions such as very deep networks highlights the difficulty of finding a good starting point in the parameter
space at random ~\citep{dauphin.2014}. On the other hand, results in Figure~\ref{fig:depthexpa}
indicate that HT can guide the student to a better initial
position in the parameter space, from which we can minimize the cost through
stochastic gradient descent.  Therefore, HT provides benefits from
an optimization point of view.
%%% Regularization
Networks trained with HT also tend to yield {\em better test
performances} than the other training methods when we fix the capacity and
number of layers.  For instance, in Figure~\ref{fig:depthexpb}, the 7-layers
network, trained with hints, obtains a $+0.7\%$ performance gain on the test
set compared to the model that does not use any hints (the
accuracy increases from $89.45\%$ to $90.1\%$). As pointed by
\cite{Erhan09}, pre-training strategies can act as regularizers. These results
suggest that HT is a stronger regularizer than KD, since it leads to
better generalization performance on the test set.
%%% Depth increases the performance and hints ease optimization
Finally, Figure~\ref{fig:depthexp} highlights that deep models have better performances
than shallower ones given a fixed computational budget.  Indeed, considering
networks that are trained with hints, an 11-layer network outperforms a 5-layer network by 
an absolute improvement of $4.11\%$ for 107M multiplications and of
$3.4\%$ for 30M multiplications. Therefore, the experiments validate
our hypothesis that given a fixed number of computations, we leverage
depth in a model to achieve faster computation and better generalization.

In summary, this experiment shows that (1) using HT, we are able to train deeper models
than with standard back-propagation and KD; and (2) given a fixed capacity, deeper models
performed better than shallower ones.

\vspace*{-1mm}
\subsection{Trade-off Between Model Performance and Efficiency}
\vspace*{-1mm}

\begin{table}
\centering
\small
\begin{tabular}{|c|c|c|c|c|c|c|}
 \hline
 \textbf{Network} & \textbf{$\#$ layers} & \textbf{$\#$ params} & \textbf{$\#$ mult} & \textbf{Acc} & \textbf{Speed-up} & \textbf{Compression rate}\\ \hline
\hline
Teacher   &  5 & $\sim$9M    & $\sim$725M & $90.18\%$ & 1 & 1\\ \hline
FitNet 1 & 11 & $\sim$250K  & $\sim$30M  & $89.01\%$ &  \textbf{13.36} & \textbf{36}\\ \hline
FitNet 2 & 11 & $\sim$862K  & $\sim$108M & $91.06\%$ & 4.64 & 10.44\\ \hline
FitNet 3 & 13 & $\sim$1.6M  & $\sim$392M & $91.10\%$ & 1.37 & 5.62\\ \hline
FitNet 4 & 19 & $\sim$2.5M  & $\sim$382M & $\mathbf{91.61\%}$ & 1.52 & 3.60\\ \hline
\end{tabular}
\captionof{table}{Accuracy/Speed Trade-off on CIFAR-10.}
\label{tab:compression}
\end{table}

%% Add a GPU version?
To evaluate FitNets efficiency, we measure their total inference
times required for processing CIFAR-10 test examples on a GPU as well as their parameter compression.
Table~\ref{tab:compression} reports both the speed-up and compression rate obtained
by various FitNets w.r.t. the teacher model along with their number of layers, capacity and accuracies. In this experiment, we retrain our FitNets on training plus validation sets as in~\citet{Goodfellow13}, for fair comparison with the teacher.

FitNet 1, our smallest network, with $36\times$ less capacity than the teacher, is one order of magnitude faster than the teacher and only witnesses a minor performance decrease of $1.3\%$. FitNet 2, slightly increasing the capacity, outperforms
the teacher by $0.9\%$, while still being faster by a strong $4.64$ factor.
By further increasing network capacity and depth in FitNets 3 and 4, we improve the performance gain, up to $1.6\%$, and still
remain faster than the teacher. Although a trade-off between speed and accuracy is introduced by the compression rate, FitNets tend to be significantly faster, matching or outperforming their teacher, even when having low capacity. 

A few works such as matrix factorization \citep{Jaderberg14,Denton14} focus on speeding-up deep networks' convolutional layers at the expense of slightly deteriorating their performance. Such approaches are complementary to FitNets and could be used to further speed-up the FitNet's convolutional layers.

Other works related to quantization schemes \citep{Chen10,Jegou11,Gong14} aim at reducing storage requirements. Unlike FitNets, such approaches witness a little decrease in performance when compressing the network parameters. Exploiting depth allows FitNets to obtain performance improvements w.r.t. their teachers, even when reducing the number of parameters $10\times$. However, we believe that quantization approaches are also complementary to FitNets and could be used to further reduce the storage requirements. It would be interesting to compare how much redundancy is present in the filters of the teacher networks w.r.t. the filters of the FitNet and, therefore, how much FitNets filters could be compressed without witnessing significant performance drop. This analysis is out of the scope of the paper and is left as future work.

\vspace*{-1mm}
\section{Conclusion}
\label{sec:concl}
\vspace*{-1mm}

We proposed a novel framework to compress \emph{wide} and \emph{deep}
networks into \emph{thin} and \emph{deeper} ones, by introducing
\emph{intermediate-level hints} from the teacher hidden layers to guide the
training process of the student. We are able to use these hints to train
very deep student models with less parameters, which can generalize better
and/or run faster than their teachers. We provided empirical evidence that hinting the inner layers of a thin and deep network with the hidden state of a teacher network generalizes better than hinting them with the classification targets. Our experiments on benchmark datasets emphasize that deep networks with low capacity are able to extract
feature representations that are comparable or even better than networks
with as much as 10 times more parameters. The hint-based training suggests
that more efforts should be devoted to explore new training strategies to
leverage the power of deep networks.

\vspace*{-1mm}
\subsubsection*{Acknowledgments}
\vspace*{-1mm}

We thank the developers of Theano
\citep{Bastien-2012} and Pylearn2 \citep{pylearn2_arxiv_2013} and the computational resources provided by Compute Canada and Calcul Qu\'ebec. This work has been partially supported by NSERC, CIFAR, and Canada Research Chairs, Project TIN2013-41751, grant 2014-SGR-221 and Spanish MINECO grant TIN2012-38187-C03.

\vspace*{-0mm}

{\small
\bibliography{iclr2015}

\begin{thebibliography}{31}
\providecommand{\natexlab}[1]{#1}
\providecommand{\url}[1]{\texttt{#1}}
\expandafter\ifx\csname urlstyle\endcsname\relax
  \providecommand{\doi}[1]{doi: #1}\else
  \providecommand{\doi}{doi: \begingroup \urlstyle{rm}\Url}\fi

\bibitem[Ba \& Caruana(2014)Ba and Caruana]{ba.2013}
Ba, J. and Caruana, R.
\newblock Do deep nets really need to be deep?
\newblock In \emph{NIPS}, pp.\  2654--2662. 2014.

\bibitem[Bastien et~al.(2012)Bastien, Lamblin, Pascanu, Bergstra, Goodfellow,
  Bergeron, Bouchard, Warde-Farley, and Bengio]{Bastien-2012}
Bastien, F., Lamblin, P., Pascanu, R., Bergstra, J., Goodfellow, I., Bergeron,
  A., Bouchard, N., Warde-Farley, D., and Bengio, Y.
\newblock {T}heano: new features and speed improvements.
\newblock Deep Learning \& Unsupervised Feature Learning Workshop, NIPS, 2012.

\bibitem[Bengio(2009)]{bengio.2009}
Bengio, Y.
\newblock Learning deep architectures for {AI}.
\newblock \emph{Foundations and trends in Machine Learning}, 2009.

\bibitem[Bengio et~al.(2007)Bengio, Lamblin, Popovici, and
  Larochelle]{Bengio07}
Bengio, Y., Lamblin, P., Popovici, D., and Larochelle, H.
\newblock Greedy layer-wise training of deep networks.
\newblock In \emph{NIPS}, pp.\  153--160, 2007.

\bibitem[Bengio et~al.(2013)Bengio, Courville, and Vincent]{bengio.2013}
Bengio, Y., Courville, A., and Vincent, P.
\newblock Representation learning: A review and new perspectives.
\newblock \emph{TPAMI}, 2013.

\bibitem[Bucila et~al.(2006)Bucila, Caruana, and Niculescu-Mizil]{Bucila06}
Bucila, C., Caruana, R., and Niculescu-Mizil, A.
\newblock Model compression.
\newblock In \emph{KDD}, pp.\  535--541, 2006.

\bibitem[Chen et~al.(2010)Chen, Guan, and Wang]{Chen10}
Chen, Yongjian, Guan, Tao, and Wang, Cheng.
\newblock Approximate nearest neighbor search by residual vector quantization.
\newblock \emph{Sensors}, 10\penalty0 (12):\penalty0 11259--11273, 2010.

\bibitem[Chen-Yu et~al.(2014)Chen-Yu, Saining, Patrick, Zhengyou, and
  Zhuowen]{Lee14}
Chen-Yu, L., Saining, X., Patrick, G., Zhengyou, Z., and Zhuowen, T.
\newblock Deeply-supervised nets.
\newblock \emph{CoRR}, abs/1409.5185, 2014.

\bibitem[Cho et~al.(2012)Cho, Raiko, Ilin, and Karhunen]{Cho12}
Cho, Kyunghyun, Raiko, Tapani, Ilin, Alexander, and Karhunen, Juha.
\newblock A two-stage pretraining algorithm for deep {Boltzmann} machines.
\newblock In \emph{{NIPS} Workshop on Deep Learning and Unsupervised Feature
  Learning}, 2012.

\bibitem[Dauphin et~al.(2014)Dauphin, Pascanu, Gulcehre, Cho, Ganguli, and
  Bengio]{dauphin.2014}
Dauphin, Y., Pascanu, R., Gulcehre, C., Cho, K., Ganguli, S., and Bengio, Y.
\newblock Identifying and attacking the saddle point problem in
  high-dimensional non-convex optimization.
\newblock In \emph{NIPS}, 2014.

\bibitem[Denton et~al.(2014)Denton, Zaremba, Bruna, LeCun, and
  Fergus]{Denton14}
Denton, Emily~L, Zaremba, Wojciech, Bruna, Joan, LeCun, Yann, and Fergus, Rob.
\newblock Exploiting linear structure within convolutional networks for
  efficient evaluation.
\newblock In \emph{NIPS}, pp.\  1269--1277. 2014.

\bibitem[Erhan et~al.(2009)Erhan, Manzagol, Bengio, Bengio, and
  Vincent]{Erhan09}
Erhan, D., Manzagol, P.A., Bengio, Y., Bengio, S., and Vincent, P.
\newblock The difficulty of training deep architectures and the effect of
  unsupervised pre-training.
\newblock In \emph{AISTATS}, pp.\  153--160, 2009.

\bibitem[Gong et~al.(2014)Gong, Liu, Yang, and Bourdev]{Gong14}
Gong, Yunchao, Liu, Liu, Yang, Min, and Bourdev, Lubomir.
\newblock Compressing deep convolutional networks using vector quantization.
\newblock \emph{CoRR}, abs/1412.6115, 2014.

\bibitem[Goodfellow et~al.(2013{\natexlab{a}})Goodfellow, Warde-Farley,
  Lamblin, Dumoulin, Mirza, Pascanu, Bergstra, Bastien, and
  Bengio]{pylearn2_arxiv_2013}
Goodfellow, I.~J., Warde-Farley, D., Lamblin, P., Dumoulin, V., Mirza, M.,
  Pascanu, R., Bergstra, J., Bastien, F., and Bengio, Y.
\newblock Pylearn2: a machine learning research library.
\newblock \emph{arXiv preprint arXiv:1308.4214}, 2013{\natexlab{a}}.

\bibitem[Goodfellow et~al.(2013{\natexlab{b}})Goodfellow, Warde-Farley, Mirza,
  Courville, and Bengio]{Goodfellow13}
Goodfellow, I.J., Warde-Farley, D., Mirza, M., Courville, A., and Bengio, Y.
\newblock Maxout networks.
\newblock In \emph{ICML}, 2013{\natexlab{b}}.

\bibitem[Gulcehre \& Bengio(2013)Gulcehre and Bengio]{Gulcehre13}
Gulcehre, C. and Bengio, Y.
\newblock Knowledge matters: Importance of prior information for optimization.
\newblock In \emph{ICLR}, 2013.

\bibitem[Hinton et~al.(2006)Hinton, Osindero, and Teh]{Hinton06_NC}
Hinton, G.~E., Osindero, S., and Teh, Y.-W.
\newblock A fast learning algorithm for deep belief nets.
\newblock \emph{Neural Computation}, 18\penalty0 (7):\penalty0 1527--1554,
  2006.

\bibitem[Hinton \& Dean(2014)Hinton and Dean]{Hinton14}
Hinton, G.~Vinyals, O. and Dean, J.
\newblock Distilling knowledge in a neural network.
\newblock In \emph{Deep Learning and Representation Learning Workshop, NIPS},
  2014.

\bibitem[Jaderberg et~al.(2014)Jaderberg, Vedaldi, and Zisserman]{Jaderberg14}
Jaderberg, M., Vedaldi, A., and Zisserman, A.
\newblock Speeding up convolutional neural networks with low rank expansions.
\newblock In \emph{BMVC}, 2014.

\bibitem[J\'egou et~al.(2011)J\'egou, Douze, and Schmid]{Jegou11}
J\'egou, Herv\'e, Douze, Matthijs, and Schmid, Cordelia.
\newblock Product quantization for nearest neighbor search.
\newblock \emph{IEEE TPAMI}, 33\penalty0 (1):\penalty0 117--128, 2011.

\bibitem[Koestinger et~al.(2011)Koestinger, Wohlhart, Roth, and Bischof]{AFLW}
Koestinger, M., Wohlhart, P., Roth, P.M., and Bischof, H.
\newblock Annotated facial landmarks in the wild: A large-scale, real-world
  database for facial landmark localization.
\newblock In \emph{First IEEE International Workshop on Benchmarking Facial
  Image Analysis Technologies}, 2011.

\bibitem[Krizhevsky \& Hinton(2009)Krizhevsky and Hinton]{Krizhevsky09}
Krizhevsky, A. and Hinton, G.
\newblock Learning multiple layers of features from tiny images.
\newblock \emph{Master's thesis, Department of Computer Science, University of
  Toronto}, 2009.

\bibitem[Larochelle et~al.(2007)Larochelle, Erhan, Courville, Bergstra, and
  Bengio]{Larochelle07}
Larochelle, H., Erhan, D., Courville, A., Bergstra, J., and Bengio, Y.
\newblock An empirical evaluation of deep architectures on problems with many
  factors of variation.
\newblock In \emph{ICML}, pp.\  473--480, 2007.

\bibitem[LeCun et~al.(1998)LeCun, Bottou, Bengio, and Haffner]{Lecun98}
LeCun, Y., Bottou, L., Bengio, Y., and Haffner, P.
\newblock Gradient-based learning applied to document recognition.
\newblock \emph{Proceedings of the IEEE}, 86\penalty0 (11):\penalty0
  2278--2324, November 1998.

\bibitem[Montufar et~al.(2014)Montufar, Pascanu, Cho, and Bengio]{Montufar14}
Montufar, G.F., Pascanu, R., Cho, K., and Bengio, Y.
\newblock On the number of linear regions of deep neural networks.
\newblock In \emph{NIPS}. 2014.

\bibitem[Netzer et~al.(2011)Netzer, Wang, Coates, Bissacco, Wu, and
  Ng]{netzer.2011}
Netzer, Y., Wang, T., Coates, A., Bissacco, A., Wu, B., and Ng, A.
\newblock Reading digits in natural images with unsupervised feature learning.
\newblock In \emph{Deep Learning \& Unsupervised Feature Learning Workshop,
  NIPS}, 2011.

\bibitem[Russakovsky et~al.(2014)Russakovsky, Deng, Su, Krause, Satheesh, Ma,
  Huang, Karpathy, Khosla, Bernstein, Berg, and Fei-Fei]{ImageNet}
Russakovsky, O., Deng, J., Su, H., Krause, J., Satheesh, S., Ma, S., Huang, Z.,
  Karpathy, A., Khosla, A., Bernstein, M., Berg, A.~C., and Fei-Fei, L.
\newblock {ImageNet Large Scale Visual Recognition Challenge}, 2014.

\bibitem[Simonyan \& Zisserman(2014)Simonyan and Zisserman]{simonyan.2014}
Simonyan, K. and Zisserman, A.
\newblock Very deep convolutional networks for large-scale image recognition.
\newblock \emph{CoRR}, abs/1409.1556, 2014.

\bibitem[Szegedy et~al.(2014)Szegedy, Liu, Jia, Sermanet, Reed, D.A., Erhan,
  Vanhoucke, and Rabinovich]{Szegedy14}
Szegedy, C., Liu, W., Jia, Y., Sermanet, P., Reed, S., D.A., Erhan, D.,
  Vanhoucke, V., and Rabinovich, A.
\newblock Going deeper with convolutions.
\newblock \emph{CoRR}, abs/1409.4842, 2014.

\bibitem[Tieleman \& Hinton(2012)Tieleman and Hinton]{Tieleman12}
Tieleman, T. and Hinton, G.
\newblock {Lecture 6.5---RmsProp: Divide the gradient by a running average of
  its recent magnitude}.
\newblock COURSERA: Neural Networks for Machine Learning, 2012.

\bibitem[Weston et~al.(2008)Weston, Ratle, and Collobert]{WestonJ2008-small}
Weston, J., Ratle, F., and Collobert, R.
\newblock Deep learning via semi-supervised embedding.
\newblock In \emph{ICML}, 2008.

\end{thebibliography}
\bibliographystyle{iclr2015}}

\newpage
\appendix

\section{Supplementary Material: Network Architectures and Training Procedures}

In the supplementary material, we describe all network architectures and hyper-parameters used throughout the paper.

\subsection{CIFAR-10/CIFAR-100}
\label{sec:cifar10}

In this section, we describe the teacher and FitNet architectures as well as hyper-parameters used in both CIFAR-10/CIFAR-100 experiments.

\subsubsection{Teachers}
We used the CIFAR-10/CIFAR-100 maxout convolutional networks reported in~\citet{Goodfellow13} as teachers. Both teachers have the same architecture, composed of 3 convolutional hidden layers of 96-192-192 units, respectively. Each convolutional layer is followed by a maxout non-linearity (with 2 linear pieces) and a max-pooling operator with respective windows sizes of 4x4, 4x4 and 2x2 pixels. All max-pooling units have an overlap of 2x2 pixels. The third convolutional layer is followed by a fully-connected maxout layer of 500 units (with 5 linear pieces) and a top softmax layer. The CIFAR-10/CIFAR-100 teachers  are trained using stochastic gradient descent and momentum. Please refer to \citet{Goodfellow13} for more details.

\subsubsection{FitNets}

Here, we describe the FitNet architectures used in the Section \ref{sec:exp} and Section \ref{sec:empres}. Each FitNet is composed of successive zero-padded convolutional layers of
kernel size $3 \times 3$, followed by a maxout non-linearity with two linear
pieces. A non-overlapping $2 \times 2$ max-pooling follows some
of the convolutional layers; each network has a total of 3 max-pooling units. The last max-pooling takes the maximum over all remaining spatial dimensions, leading to a $1 \times 1$ vector representation. The last convolutional layer is followed by a fully-connected and a softmax layer, as the ones on CIFAR-10/100 teachers.

Table~\ref{tab:cifar100_depthfitnet} describes the architectures used for the
depth experiment in Figure \ref{fig:depthexp}. Table~\ref{tab:cifar100_fitnet} describes the architectures for the efficiency-performance trade-off experiment in Table \ref{tab:compression}. The results reported in Table \ref{tab:cifar10}, Table \ref{tab:cifar100} and Table \ref{tab:svhn} correspond to the FitNet 4 architecture.

All FitNet parameters were initialized randomly in U(-0.005,0.005). We used stochastic gradient descent with RMSProp \citep{Tieleman12} to train the FitNets, with an initial
learning rate $0.005$ and a mini-batch size of 128. Parameter $\lambda$ in Eq. (\ref{eq:toplayer})
was initialized to $4$ and decayed linearly during $500$ epochs reaching $\lambda = 1$. The relaxation term $\tau$ was set to 3.

On CIFAR-10, we divided the training set into 40K training examples and
10K validation examples. We trained stage 1 by minimizing Eq.~(\ref{eq:hints})
and stopped the training after 100 epochs of no validation error improvement, performing a maximum of 500 epochs.  After that, we trained stage 2 by minimizing Eq. (\ref{eq:toplayer}) using RMSprop, the same stopping criterion and the same hyper-parameters as stage 1. We picked the optimal number of epochs according to the above-mentioned stopping criterion and retrained the FitNet on the
whole 50K training examples (training + validation sets).  

On CIFAR-100, we trained directly on the whole training set using stochastic gradient descent with RMSprop, the same hyper-parameters as CIFAR-10 FitNets and the number of epochs determined by CIFAR-10 stopping criterion.

\begin{table}
\center
\begin{tabular}{|c|c|c|c|}
\hline
 5 Layer & 7 Layer & 9 Layer & 11 Layer\\
\hline
\hline
 conv 3x3x64 (3x3x128) &  conv 3x3x16 (3x3x32)  & conv 3x3x16 (3x3x32)  & conv 3x3x16 (3x3x16)\\
 pool 2x2          &  conv 3x3x32 (3x3x64)  & conv 3x3x32 (3x3x32)  & conv 3x3x16 (3x3x32)\\
                   &  pool 2x2          & pool 2x2          & conv 3x3x16 (3x3x32)\\
                   &                    &                   & pool 2x2\\
\hline
 conv 3x3x64 (3x3x128) &  conv 3x3x32 (3x3x80)  & conv 3x3x32 (3x3x64)  & conv 3x3x32 (3x3x48)\\
 pool 2x2          &  conv 3x3x64 (3x3x80)  & conv 3x3x32 (3x3x80)  & conv 3x3x32 (3x3x64)\\
                   &  pool 2x2          & conv 3x3x32 (3x3x80)  & conv 3x3x32 (3x3x80)\\
                   &                    & pool 2x2          & pool 2x2\\
\hline
 conv 3x3x64 (3x3x128) &  conv 3x3x64 (3x3x128) & conv 3x3x48 (3x3x96)  & conv 3x3x48 (3x3x96)\\
 pool 8x8          &  pool 8x8          & conv 3x3x64 (3x3x128) & conv 3x3x48 (3x3x96)\\
                   &                    & pool 8x8          & conv 3x3x64 (3x3x128)\\
                   &                    &                   & pool 8x8\\
\hline
 fc  & fc & fc & fc \\
 softmax  & softmax & softmax & softmax\\
\hline
hint: 2$\leftarrow$2 & hint: 4$\leftarrow$2 & hint: 5$\leftarrow$2 & hint: 7$\leftarrow$2\\
\hline
\end{tabular}

\caption{Fitnet architectures with a computational budget of  30M  (or 107M) of multiplications: conv $s_x \times s_y \times  c$ is a convolution of kernel size $s_x \times  s_y$ with $c$ outputs channels; pool $s_x \times s_y$ is a non-overlapping
pooling of size $s_x \times s_y$; fc stands for fully connected.
 hint: FitNet $\leftarrow$ teacher specifies
the hint and guided layers used for hint-based training, respectively.}
\label{tab:cifar100_depthfitnet}
\end{table}

\begin{table}
\center
\begin{tabular}{|c|c|c|c|}
\hline
 FitNet 1 & FitNet 2 & FitNet 3 & FitNet 4\\
\hline
\hline
 conv 3x3x16 &  conv 3x3x16 & conv 3x3x32 & conv 3x3x32\\
 conv 3x3x16 &  conv 3x3x32 & conv 3x3x48 & conv 3x3x32\\
 conv 3x3x16 &  conv 3x3x32 & conv 3x3x64 & conv 3x3x32\\
 pool 2x2    &  pool 2x2    & conv 3x3x64 & conv 3x3x48\\
             &              & pool 2x2    & conv 3x3x48\\
             &              &             & pool 2x2\\
\hline
 conv 3x3x32 &  conv 3x3x48 & conv 3x3x80 & conv 3x3x80\\
 conv 3x3x32 &  conv 3x3x64 & conv 3x3x80 & conv 3x3x80\\
 conv 3x3x32 &  conv 3x3x80 & conv 3x3x80 & conv 3x3x80\\
 pool 2x2    &  pool 2x2    & conv 3x3x80 & conv 3x3x80\\
             &              & pool 2x2    & conv 3x3x80\\
             &              &             & conv 3x3x80\\
             &              &             & pool 2x2\\
\hline
 conv 3x3x48 & conv 3x3x96   & conv 3x3x128 & conv 3x3x128\\
 conv 3x3x48 & conv 3x3x96   & conv 3x3x128 & conv 3x3x128\\
 conv 3x3x64 & conv 3x3x128 & conv 3x3x128 & conv 3x3x128\\
 pool 8x8        & pool 8x8           & pool 8x8           & conv 3x3x128\\
                         &                            &                            & conv 3x3x128\\
                         &                            &                            & conv 3x3x128\\
                         &                            &                            & pool 8x8\\
\hline
 fc   & fc & fc & fc\\
 softmax  & softmax & softmax & softmax\\
\hline
hint: 6$\leftarrow$2 & hint: 6$\leftarrow$2 & hint: 8$\leftarrow$2 & hint: 11$\leftarrow$2\\
\hline
\end{tabular}

\caption{Performance-Efficiency FitNet architectures.}
\label{tab:cifar100_fitnet}
\end{table}

\subsection{MNIST}

In this section, we describe the teacher and FitNet architectures as well as the hyper-parameters used in the MNIST experiments.

%% About exp
We trained a teacher network of maxout convolutional layers as reported in
\citet{Goodfellow13}. The teacher architecture has three convolutional maxout hidden
layers (with 2 linear pieces each) of 48-48-24 units, respectively, followed by a spatial max-pooling of
4x4-4x4-2x2 pixels, with an overlap of 2x2 pixels. The 3rd hidden layer is
followed by a fully-connected softmax layer. As is \citet{Goodfellow13}, we
added zero padding to the second convolutional layer. 

We designed a FitNet twice as deep as the teacher network and with roughly $8\%$ of the parameters. The student architecture has 6 maxout convolutional hidden layers (with 2 linear pieces each) of 16-16-16-16-12-12 units, respectively. Max-pooling is only applied every second layer in regions of 4x4-4x4-2x2 pixels, with an overlap of 2x2
pixels. The 6th convolutional hidden layer is followed by a fully-connected
softmax layer.

The teacher network was trained as described in \citet{Goodfellow13}. The FitNet was trained in a stage-wise fashion as described in Section \ref{sec:method}. We divided the training set into a training set of 50K samples and a validation set of 10K samples. 

All network parameters where initialized randomly in U(-0.005,0.005). In the first stage, the 4th layer of the FitNet was trained to mimic the 2nd layer of the teacher network, by minimizing Eq. (\ref{eq:hints}) through stochastic gradient descent. We used a mini-batch size of 128 samples and fixed the learning rate to 0.0005. We initialized $\lambda$ to 4 and decayed it for the first 150 epochs until it reached 1. The training was stopped according to the following criterion: after 100 epochs of no validation error improvement and performning a maximum of 500 epochs. We used the same mini-batch size, learning rate and stopping criterion to train the second stage. The relaxation term $\tau$ was set to 3.

\subsection{SVHN}
\label{sec:svhn}

In this section, we describe the teacher and FitNet architectures as well as the hyper-parameters used in the SVHN experiments.

We used SVHN maxout convolutional network described in as \citet{Goodfellow13} teacher. The network is composed of 3 convolutional hidden layers of 64-128-128 units, respectively, followed by a fully-connected maxout layer of 400 units and a top softmax layer. The teacher training was carried out as in~\citet{Goodfellow13}. 

We used the FitNet 4 architecture outlined in Table~\ref{tab:cifar100_fitnet}, initializing the network parameters randomly in U(-0.005,0.005) and training with the same hyper-parameters as in CIFAR-10. In this case, we used the same early stopping as in CIFAR-10, but we did not retrain the FitNet on the whole training set (training + validation). The same hyper-parameters where used for both stages.

\subsection{AFLW}

In this section, we describe the teacher and FitNet architectures as well as the hyper-parameters used in the AFLW experiments. 

We trained a teacher network of 3 ReLU convolutional layers of 128-512-512 units, respectively, followed by a sigmoid layer. Non-overlapping max-pooling of size $2 \times 2$ was performed after the first convolutional layer. We used receptive fields of 3-2-5 for each layer, respectively.

We designed two FitNets of 7 ReLU convolutional layers. Fitnet 1's layers have 16-32-32-32-32-32-32-32 units, respectively, followed by a sigmoid layer. Fitnet 2's layers have 32-64-64-64-64-64-64-64 units, respectively, followed by a sigmoid layer. In both cases, we used receptive fields of $3 \times 3$ and, due to the really small image resolution, we did not perform any max-pooling. 

All network parameters of both FitNets where initialized randomly in U(-0.05,0.05). Both FitNets were trained in the stage-wise fashion described in Section \ref{sec:method}. We used $90\%$ of the data for training. In the first stage, the 5th layer of the FitNets were trained to mimic the 3rd layer of the teacher network, by minimizing Eq. (\ref{eq:hints}) through stochastic gradient descent. We used a mini-batch size of 128 samples and initialized the learning rate to 0.001 and decayed it for the first 100 epochs until reaching 0.01. We also used momentum. We initialized momentum to 0.1 and saturated it to 0.9 at epoch 100. We picked the best validation value after a 500 epochs. We used the same mini-batch size, learning rate and stopping criterion to train the second stage. The relaxation term $\tau$ was set to 3.

\end{document}